\documentclass[letterpaper, 10 pt, conference]{ieeeconf}  % Comment this line out if you need a4paper

\IEEEoverridecommandlockouts                              % This command is only needed if 
\usepackage{graphicx} % for pdf, bitmapped graphics files
\usepackage{epsfig} % for postscript graphics files
\usepackage{times} % assumes new font selection scheme installed
\usepackage{subcaption}
\usepackage{amsmath} % assumes amsmath package installed
\usepackage{amssymb}  % assumes amsmath package installed
\usepackage{xspace}

\newcommand{\PAR}[1]{\vskip2pt \noindent {\bf #1~}}

\makeatletter
\DeclareRobustCommand\onedot{\futurelet\@let@token\@onedot}
\def\@onedot{\ifx\@let@token.\else.\null\fi\xspace}

\def\eg{\emph{e.g}\onedot} 
\def\ie{\emph{i.e}\onedot}

\def\etal{\emph{et al}\onedot}
\makeatother

\title{\LARGE \bf
Single-Shot Panoptic Segmentation
}

\author{Mark Weber$^{1}$, Jonathon Luiten$^{1}$ and Bastian Leibe$^{1}$% <-this % stops a space
\thanks{$^{1}$RWTH Aachen University \newline
        \indent{\tt\small mark.weber1@rwth-aachen.de}}%
}

\begin{document}

\maketitle
\thispagestyle{empty}
\pagestyle{empty}

%%%%%%%%%%%%%%%%%%%%%%%%%%%%%%%%%%%%%%%%%%%%%%%%%%%%%%%%%%%%%%%%%%%%%%%%%%%%%%%%
\begin{abstract}
We present a novel end-to-end single-shot method that segments countable object instances (\texttt{things}) as well as background regions (\texttt{stuff}) into a non-overlapping panoptic segmentation at almost video frame rate. Current state-of-the-art methods are far from reaching video frame rate and mostly rely on merging instance segmentation with semantic background segmentation, making them impractical to use in many applications such as robotics. Our approach relaxes this requirement by using an object detector but is still able to resolve \textit{inter-} and \textit{intra-class} overlaps to achieve a non-overlapping segmentation. On top of a shared encoder-decoder backbone, we utilize multiple branches for semantic segmentation, object detection, and instance center prediction. Finally, our panoptic head combines all outputs into a panoptic segmentation and can even handle conflicting predictions between branches as well as certain false predictions. Our network achieves 32.6\% PQ on MS-COCO at 23.5 FPS, opening up panoptic segmentation to a broader field of applications.
\end{abstract}

%%%%%%%%%%%%%%%%%%%%%%%%%%%%%%%%%%%%%%%%%%%%%%%%%%%%%%%%%%%%%%%%%%%%%%%%%%%%%%%%
\section{Introduction}

Scene understanding is one of the fundamental yet challenging tasks of computer vision. Traditional scene understanding tasks include object detection, instance segmentation, and semantic segmentation. The former tasks address classification and localization of countable foreground objects, while the latter focuses on segmenting image regions belonging to the same class without distinguishing different instances. Both tasks are necessary for a complete scene understanding. However, in recent years, the approaches used for these tasks have been quite different. For example, region-proposal networks \cite{He_17_ICCV} are used for object detection and instance segmentation, while fully convolutional networks \cite{Zhao_17_CVPR} have been used for semantic segmentation.

\begin{figure}
    \centering
    \begin{subfigure}[t]{0.49\linewidth}
        \includegraphics[width=1.00\linewidth]{images/input/000000482800.jpg}
        \caption{Input}
    \end{subfigure}
    \begin{subfigure}[t]{0.49\linewidth}
        \includegraphics[width=1.00\linewidth]{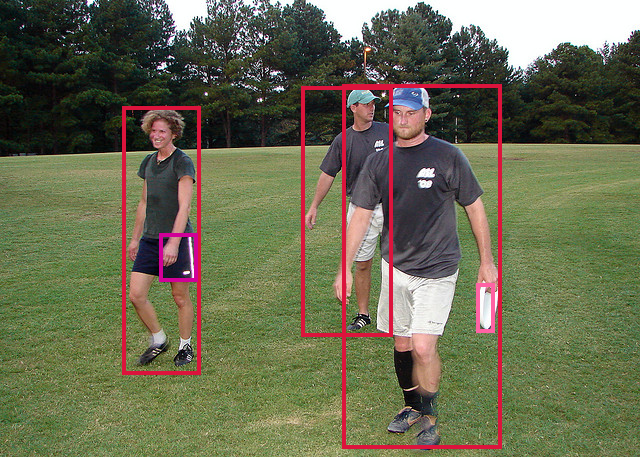}
        \caption{Object Detection}
    \end{subfigure}
    \begin{subfigure}[t]{0.49\linewidth}
        \includegraphics[width=1.00\linewidth]{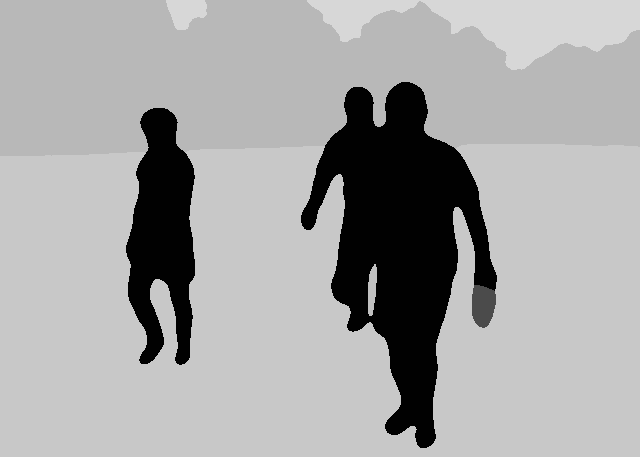}
        \caption{Semantic Segmentation}
    \end{subfigure}
    \begin{subfigure}[t]{0.49\linewidth}
        \includegraphics[width=1.00\linewidth]{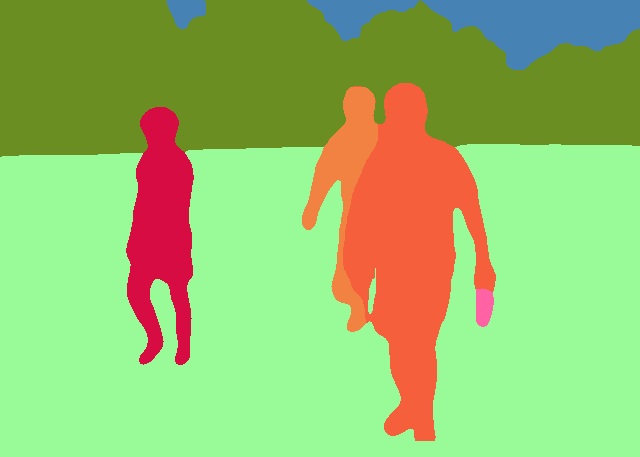}
        \caption{Panoptic Segmentation}
    \end{subfigure}
    \caption{Our single-shot panoptic segmentation network generates object detection and semantic segmentation results that our panoptic head combines to panoptic segmentation. Remarkably, our panoptic head can resolve inter- and intra-class overlaps and even handle wrong detections, \eg, the detected handbag (top-right image).}
	\label{fig:teaser}
\end{figure}

Panoptic segmentation (Fig.~\ref{fig:teaser}) was introduced by Kirillov \etal to tackle the problem of diverging modeling strategies for different tasks in the field of visual scene understanding \cite{Kirillov_19_CVPR_PanSeg}. Current methods for this task are mostly two-stage approaches which extend the proposal-based Mask R-CNN \cite{He_17_ICCV}. Similarly to object detection and instance segmentation, these approaches achieve high accuracy but suffer from slow run-times that make them infeasible for many real-world applications. In contrast, single-shot methods have the potential to be faster and simpler, enabling many real-world applications with substantial efficiency requirements such as robotics and autonomous driving. For many such applications, the desire to navigate requires some scene understanding. The better this scene understanding is, the better the planning and navigation task becomes. So far, single-shot approaches for object detection like YOLO \cite{Redmon_16_CVPR}, SSD \cite{Liu_16_ECCV}, and their successors have already made substantial impact in robotics. Hence, our goal is to open up panoptic segmentation to the robotics field and we argue it is important to not just push for algorithms with the absolute best performance, but also to explore the trade-off between speed and accuracy.

% Teaser image

In this paper, we present a unified single-shot network to tackle the task of panoptic segmentation. Unlike previous methods that maximize only for accuracy, we base our model design choices on the speed-accuracy trade-off, which is of special interest for the robotics community. We use conceptionally simple, yet efficient components for the tasks of object detection and semantic segmentation. In contrast to many related methods that heuristically merge outputs from different branches, we investigate more conflict-resolving merging procedures. We present a novel panoptic head that uses instance center predictions with object detection and semantic segmentation to extend on naive merging. This novel approach can resolve both intra- and inter-class overlaps to obtain instance segmentation and is able correct false predictions of individual components. We extensively evaluate our method on the popular COCO dataset \cite{Lin_14_ECCV} and demonstrate the effectiveness of our approach. In particular, we achieve huge run-time gains compared to state-of-the-art techniques. Furthermore, we compare our method also to one concurrent work on single-shot panoptic segmentation and achieve a far better speed-accuracy trade-off.

%-------------------------------------------------------------------------
\section{Related Work}
\PAR{Semantic Segmentation:} State-of-the-art methods for semantic segmentation utilize a fully convolutional neural network (FCN) \cite{Long_15_CVPR} and follow an encoder-decoder scheme. Several approaches incorporate contextual information by aggregating multi-scale features.
%Since context is of high importance for semantic segmentation, a significant improvement was proposed for the DeepLab architecture \cite{Chen_15_ICLR} and its successors \cite{Chen_16_arXiv, Chen_17_arXiv, Chen_18_ECCV}. Dilated convolutions offer a broader context without increasing the number of model parameters.
The use of dilated convolutions \cite{Chen_15_ICLR, Chen_16_arXiv, Chen_17_arXiv, Chen_18_ECCV} has allowed methods to importantly exploit a larger image context without increasing the number of model parameters.
Recently, methods using variants of pyramid pooling became dominant. PSPNet \cite{Zhao_17_CVPR} uses spatial pyramid pooling, while DeepLab proposes an atrous spatial pyramid pooling module (ASPP) \cite{Chen_18_ECCV}. With this multi-scale contextual information, these methods demonstrate high accuracy on standard benchmarks. A fully convolutional component of our network is dedicated to predicting semantic segmentation by leveraging multi-scale features from a pyramid but refrains from using dilated convolutions since they are computationally more expensive.

\PAR{Object Detection and Instance Segmentation:} In contrast to semantic segmentation, methods for object detection and instance segmentation identify different instances of the same class. Two-stage methods using region proposals have become the dominant concept for high accuracy object detection and instance segmentation, \eg, Mask R-CNN \cite{He_17_ICCV}. In these approaches a set of proposals is generated and then processed by a sub-network for classification and segmentation. While these methods achieve state-of-the-art accuracy, they are infeasible to use for many applications due to their computational costs. For object detection, single-shot methods address the speed-accuracy trade-off and achieve remarkable results. Besides YOLO \cite{Redmon_16_CVPR} and SSD \cite{Liu_16_ECCV}, RetinaNet \cite{Lin_17_ICCV} uses a dense field of anchor boxes and CornerNet \cite{Law_18_ECCV} generates heatmaps to produce bounding boxes.

While single-shot detectors are closing the accuracy gap to two-stage detectors, single-shot instance segmentation methods lag behind state-of-the-art methods. Several methods learn embeddings which are clustered to produce instances \cite{Kong_18_CVPR, DeBrabandere_17_arXiv, fathi_17_arXiv}. While this concept is straightforward, current techniques still achieve low accuracy and are not necessarily faster than two-stage methods. A different way of producing instances is to predict instance centers per pixel and to combine these with predicted bounding boxes to instances \cite{Uhrig_18_IV}. However, this technique is far from achieving state-of-the-art accuracy. Since single-shot instance segmentation methods still gain much lower accuracy than their two-stage competitors, we choose to use a single-shot object detection network (RetinaNet \cite{Lin_17_ICCV}) instead that predicts object bounding boxes. 
%in a single-shot fashion.

\PAR{Panoptic Segmentation:} Panoptic segmentation can be understood as a multi-task learning problem. Prior work on multi-task learning focuses on learning multiple tasks with a shared backbone \cite{Kendall_18_CVPR, Pfeiifer_19_GCPR}. However, conflicting predictions of multiple branches have typically not been resolved. Panoptic segmentation, as defined by~\cite{Kirillov_19_CVPR_PanSeg}, expects unique predictions for each pixel, which requires assigning one semantic class per pixel and an instance ID for all pixels belonging to a \texttt{thing} class. Since the introduction of panoptic segmentation, all published supervised methods have focused on a two-stage architecture \cite{Xiong_19_CVPR, Kirillov_19_CVPR_PanFPN, Li_19_CVPR, Liu_19_CVPR, Sofiiuk_19_arXiv}. The same holds for all high scoring entries in the COCO panoptic challenge \cite{Challenge_18_ECCV}.

The typical network design extends Mask R-CNN by adding a parallel semantic segmentation branch and a fusion step to combine instance and semantic segmentation to produce panoptic segmentation \cite{Kirillov_19_CVPR_PanFPN}. This fusion step needs to solve two major problems. First, the predicted instances by Mask R-CNN can overlap; and second, the output of both branches can be conflicting, \eg, a pixel may be predicted to belong to a car instance by one branch and to the sky by the other branch. \cite{Kirillov_19_CVPR_PanSeg, Kirillov_19_CVPR_PanFPN, Li_19_CVPR} all tackle the first problem by sorting the instances based on the confidence score above a certain threshold and defining an IoU threshold to skip instances that overlap with other instances too much. \cite{Liu_19_CVPR} propose a module that learns the spatial relationship between different semantic \textit{classes} or \textit{instances}, respectively. On top of the architecture \cite{Kirillov_19_CVPR_PanFPN}, \cite{Li_19_CVPR} adds attention layers to reduce conflicts between branches. All these approaches address the second problem merely by always taking the Mask R-CNN output and they only use the semantic segmentation for the remaining unassigned pixels. The only more sophisticated method is UPSNet \cite{Xiong_19_CVPR} that leverages a panoptic head to combine semantic segmentation and instance segmentation predictions into instance-aware logits.

Since all of these approaches are based on Mask R-CNN \cite{He_17_ICCV}, they all suffer from high computational cost and hence, cannot be used for applications that require real-time operation. Please note, that we compare our approach only to published peer-reviewed methods. To the best of our knowledge, there is no peer-reviewed published single-shot method for panoptic segmentation. Therefore, we follow the practice of \cite{Sofiiuk_19_arXiv} and also compare to concurrent work 
%\cite{Lazarow_19_arXiv, Li_18_arXiv, Yang_19_arXiv, DeGeus_19_arXiv, Sofiiuk_19_arXiv}. 
DeeperLab \cite{Yang_19_arXiv}. DeeperLab is a single-shot architecture that is based on DeepLab \cite{Chen_18_ECCV} and PersonLab \cite{Papandreou_18_ECCV}. Class-agnostic instances are predicted with the help of keypoint heatmaps. Their semantic class is determined with a majority vote from the semantic prediction. Predicted \texttt{stuff} classes are merely kept for remaining pixels.

In this work, we propose a single-shot architecture and thus, do not extend Mask R-CNN. Unlike most state-of-the-art approaches, we utilize a conflict-resolving merging head in our architecture that tackles the problem of conflicting predictions. In contrast to UPSNet \cite{Xiong_19_CVPR}, our merging head is capable of working with bounding boxes instead of instance segments, which opens our framework to a broad range of available detectors. Unlike DeeperLab \cite{Yang_19_arXiv}, we show that we are capable of achieving close to video frame rate performance for our end-to-end network.

%-------------------------------------------------------------------------
\section{Method}

\begin{figure}
	\centering
		\includegraphics[width=1.00\linewidth]{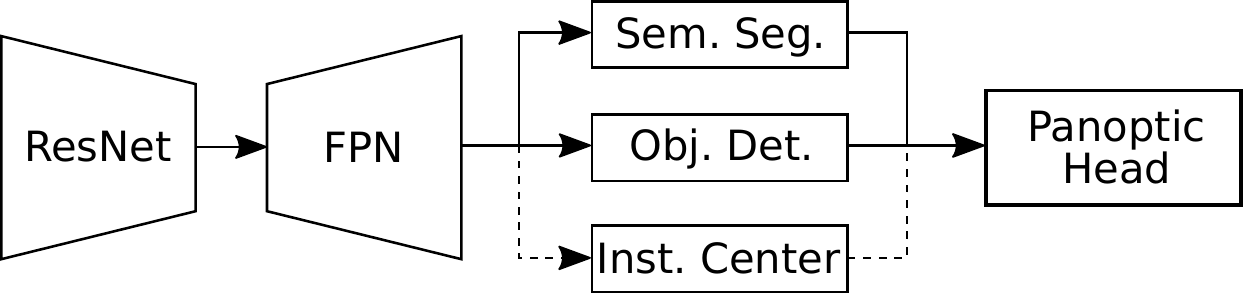}
	\caption{Our proposed network architecture leverages an encoder-decoder backbone, up to three branches and a panoptic head to produce panoptic segmentation.}
	\label{fig:architecture}
\end{figure}

Our proposed end-to-end network consists of a shared convolutional backbone that extracts multi-scale features in an encoder-decoder fashion. On top of this, we add multiple branches for the tasks of semantic segmentation, object detection, and optionally instance center prediction. All these predictions are input to our final panoptic head that produces panoptic segmentation without any post-processing merging steps involved. The overall network architecture is shown in Figure~\ref{fig:architecture}. Our network design goals are accuracy, speed, simplicity and comparability.

\subsection{Proposed Architecture}

\PAR{Backbone:} Our backbone uses a residual network (ResNet) \cite{He_16_CVPR} as the encoder and a feature pyramid network (FPN) \cite{Lin_17_CVPR} as the decoder. Hence, it offers multi-scale contextual information that can be used for both semantic segmentation as well as for object detection. We choose a standard ResNet-50 FPN to make comparisons to previous methods as fair as possible. Our shared backbone generates pyramid levels with scales from 1/128 to 1/4 resolution with each level having 256 feature dimensions.

\begin{figure}
	\centering
	\includegraphics[width=1.00\linewidth]{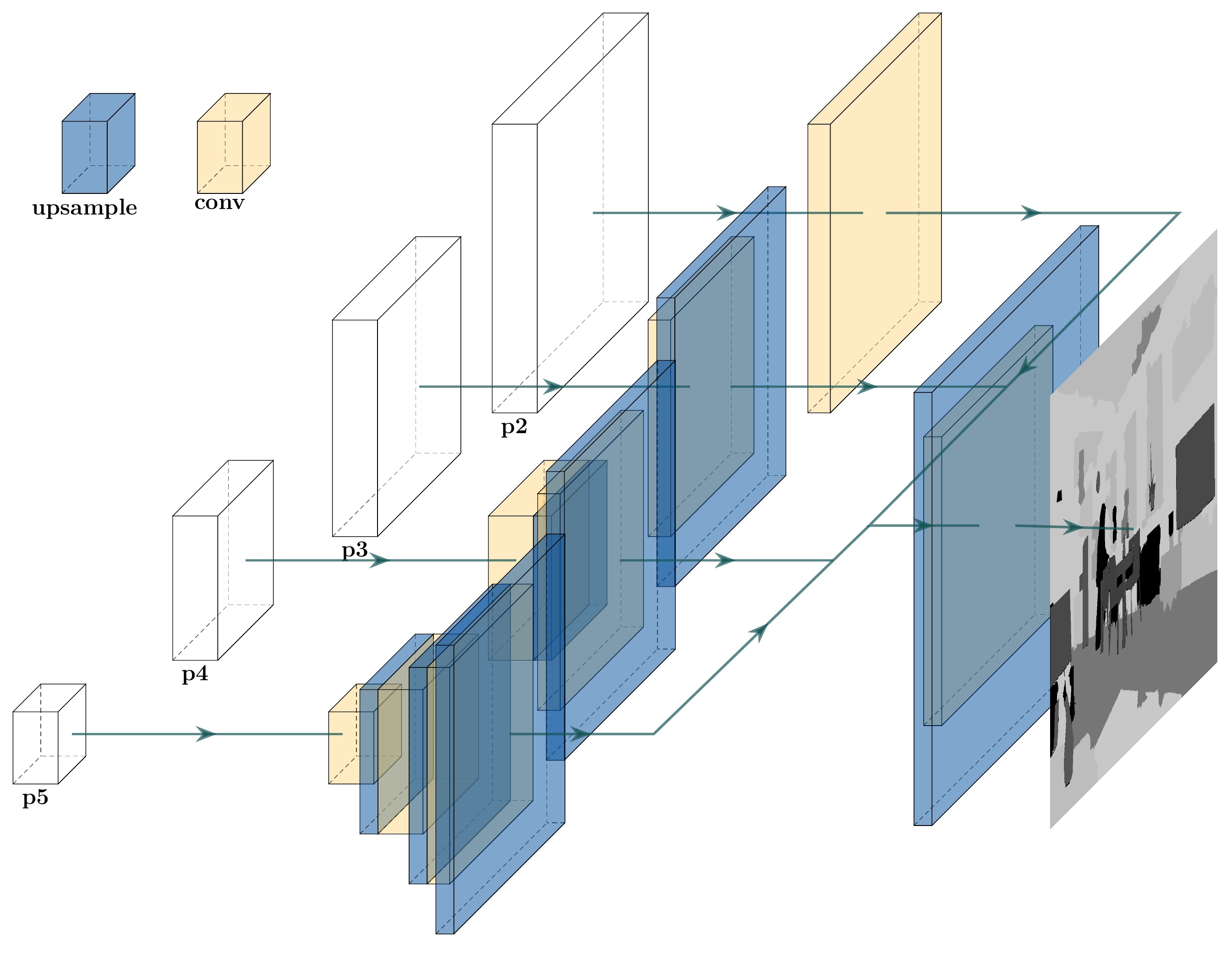}
	\caption{The semantic segmentation branch leverages the levels p2 through p5 to produce logits. Each feature pyramid tensor is convolved (yellow) and upsampled (blue) to the same spatial resolution and finally element-wise summed up before the logits are computed and upsampled to the original resolution.}
	\label{fig:sem-seg}
\end{figure}

\PAR{Semantic Segmentation Branch:} Similar to \cite{Kirillov_19_CVPR_PanFPN}, we use a lightweight fully convolutional branch to predict semantic segmentation. We take the feature pyramid levels p5 to p2, which correspond to 1/32 to 1/4 resolution, and alternate convolutions and upsampling operations until all pyramid levels are on a quarter resolution. After convolutions, we use GroupNorm layers \cite{Wu_18_ECCV} for normalization and ReLU activation functions. Finally, we aggregate all upsampled pyramid levels by accumulating them and using an additional convolution to produce semantic logits. The network is sketched in Figure~\ref{fig:sem-seg}. In most previous work, the implementation of the semantic segmentation branch varies only slightly \cite{Liu_19_CVPR, Xiong_19_CVPR, Li_19_CVPR, Sofiiuk_19_arXiv}. Since all of them achieve similar performance, we chose this lightweight branch for our purpose. However, unlike most related work, our semantic segmentation branch predicts all \texttt{things} and \texttt{stuff} classes, which is required for our panoptic head.

\begin{figure}
	\centering
	\includegraphics[width=1.00\linewidth]{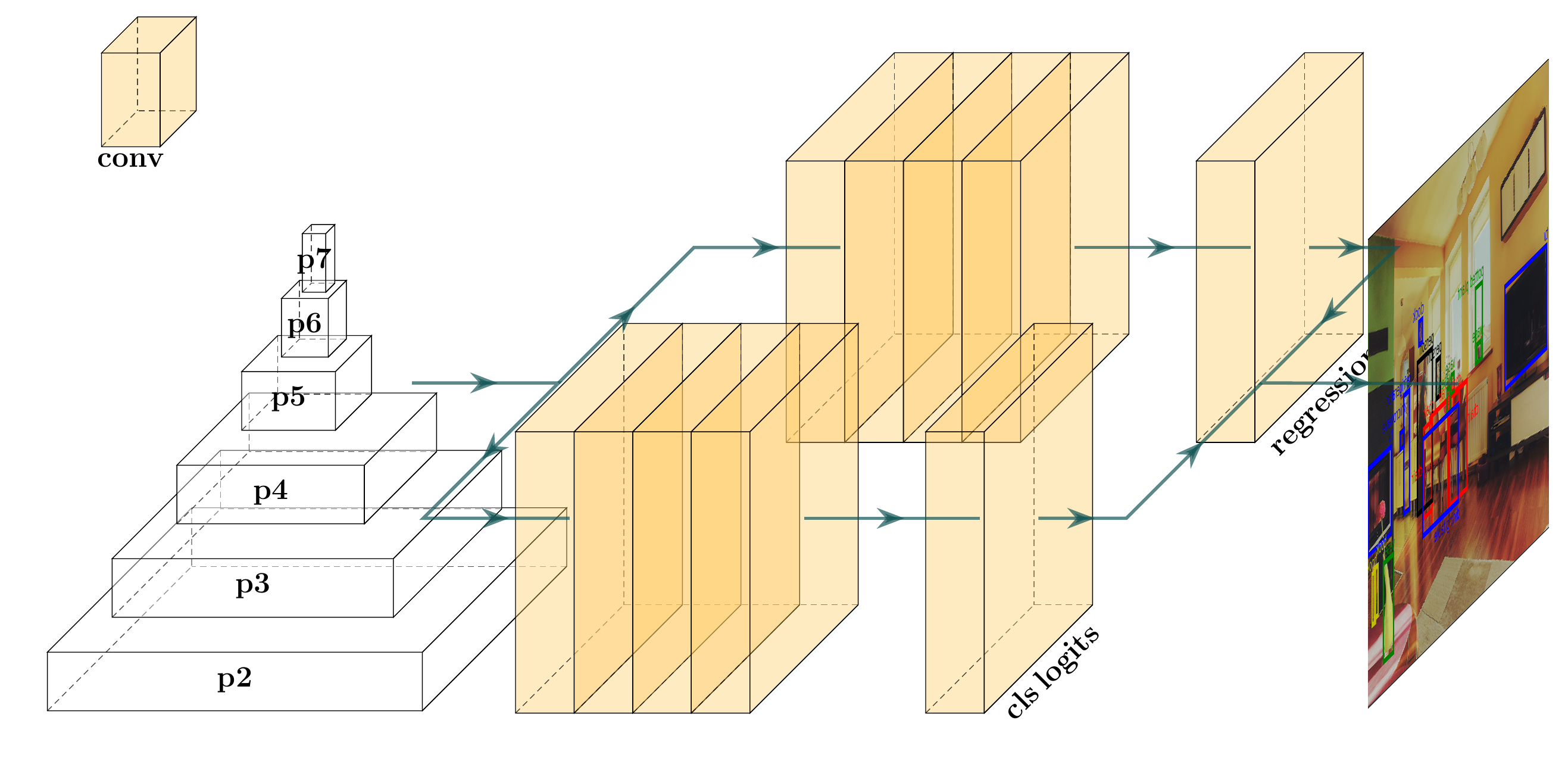}
	\caption{The object detection branch utilizes all levels of the feature pyramid to predict bounding boxes and corresponding classes. Each  pyramid level is fed through 4 convolutions (yellow) for each task. Finally, offsets for all anchor boxes and corresponding logits are predicted.}
	\label{fig:obj-det}
\end{figure}

\PAR{Object Detection Branch:} While our method is not limited to one specific object detector, we use the RetinaNet \cite{Lin_17_ICCV} architecture for demonstration purpose as it exhibits a good speed-accuracy trade-off. RetinaNet contains two fully convolutional subnetworks that classify and regress anchor boxes. The same network weights are used throughout all pyramid levels from 1/128 to 1/4 to detect \texttt{things} of all sizes. Figure~\ref{fig:obj-det} shows the object detection branch. We keep the standard hyperparameters to demonstrate the out-of-the-box performance of our overall network.

\PAR{Panoptic Head:} Taking inspiration from UPSNet \cite{Xiong_19_CVPR}, we propose a novel parameter-free panoptic head that is capable of working with object detections instead of instance logits. Our goal is to create instance-aware logits $Y$ than can be inferred like semantic logits but offer additional instance information. Therefore, we split our semantic logits $X$ into \texttt{stuff} $X_{\text{stuff}}$ and \texttt{things} $X_{\text{things}}$ logits. We merely copy $X_{\text{stuff}}$ into $Y$ as they do not need an instance ID. For $X_{\text{things}}$, we use the object detection results, which include a class $c$ and a bounding box $b$ per object. We use $c$ as the index to select the corresponding slice in $X_{\text{things}}$. In these 2D logits $X_{c}$, we use the bounding box $b$ to crop the logits, \ie, we filter out all logit values outside $b$.  We take these logits $X_{c,b}$ and stack them with the \texttt{stuff} logits $X_{\text{stuff}}$. After this selection, $X_{c,b}$ will still have the same width and height as the semantic logits. However, the number of channels in $Y$ directly depends on the number of detected objects and therefore, varies for each input.

In order to illustrate this point, consider the following example. Suppose three cars have been detected. Our network then selects
%For three detected cars, we select
the car logit slice $X_{c=\text{car}}$ three times, crops the corresponding bounding boxes $b_1, b_2, b_3$ and stacks them with $X_{\text{stuff}}$ to gain instance-aware logits $Y$. Hence, we end up with three new logit slices, each corresponding to a car instance. Regions that are segmented as \texttt{things} by the semantic segmentation branch, but that were not detected by the object detector are discarded. Moreover, falsely detected objects might still get low logit values from the semantic segmentation branch and therefore do not show up in the final panoptic segmentation. Thus, false predictions in one branch can be corrected by the other branch, which we consider a significant improvement over simple merging by copying.

Since the original logits remain, an argmax operation is still capable of achieving precise contours of objects, even if the detector does not provide such information. The only problem that arises comes from overlapping bounding boxes with the same class, while overlaps of different classes are not an issue. Since the same logit slice is selected in such cases, the overlapping regions will have the same value in our final result. Hence, an argmax operation cannot choose the correct instance. To overcome this problem, we propose and investigate different strategies.

\subsection{Overlap Resolution}

All previously published work requires instance segmentation results to produce panoptic segmentation. Part of our work is the relaxation of this requirement to object detectors. While the panoptic head addresses \textit{inter-class} overlaps, \textit{intra-class} overlaps cannot be resolved directly. Hence, we propose three different policies to overcome this shortcoming. Combined with the panoptic head, we can solve both kinds of overlaps, thus making the use of instance segmentation subnetworks optional.

\PAR{Highest-Confidence Policy:} This policy is similar to the way previous work handles overlapping instances (of any class) predicted by Mask R-CNN. Assuming the confidence score correlates strongly to the probability of the existence of an object, sorting the overlapping bounding boxes in decreasing order of confidence leads to a higher score. With this policy, very likely predictions could hide false positives.

\PAR{Smallest-First Policy:} The PQ metric \cite{Kirillov_19_CVPR_PanSeg}, used to measure the quality of panoptic segmentation, puts much emphasis on small instances since even the smallest ones count as much towards the score as the largest segments. Hence, sorting overlapping instances in increasing order of size prevents large instances from overshadowing smaller ones.

\PAR{Closest-Center Policy:} Both policies described above suffer from the effect that the overlap will be assigned to only one instance. Hence, they   introduce straight contours with sharp corners originating from the bounding boxes. While this issue might not be reflected much in the final score since the number of \textit{intra-class} bounding box overlaps might be somewhat limited, the visual quality of the result for these cases will suffer tremendously. Therefore, we propose a third branch in our framework. The task of this third branch is class-agnostic instance center prediction, which we specify as the prediction of the center of the bounding box.

Uhrig et al. \cite{Uhrig_18_IV} used instance center predictions to cluster instances directly with limited success. However, clustering predictions to achieve instance segmentation has to solve \textit{inter-} and \textit{intra-class} overlaps, while our method only needs to resolve \textit{intra-class} overlaps. \textit{Intra-class} overlaps, intuitively, formulate only a small amount of all potential overlaps. We utilize the same lightweight architecture as for semantic segmentation but predict only the absolute offset in x- and y-direction from the pixel's location. The overlap is resolved by adding the predicted offset to the pixels' location and computing the L2 distance to all the centers of boxes containing that pixel. The smallest distance gives the most likely instance for each pixel. This policy enables accurate contours assuming good instance center predictions.

\subsection{Inference and Training}

\PAR{Unknown Predictions:} While the task of panoptic segmentation is defined as assigning a semantic class and for \texttt{things} also an instance ID to every pixel, previous work makes use of an \texttt{unknown} prediction class in some cases. For example, \cite{Kirillov_19_CVPR_PanFPN, Xiong_19_CVPR} remove \texttt{stuff} regions that occupy fewer pixels than a certain threshold. Moreover, they make \texttt{unknown} predictions for certain \texttt{stuff} areas that their branches produce conflicting assignments. While this practice is debatable as it weakens the task requirements, it is encouraged by improved scores in the PQ metric. For some pixels, the semantic segmentation branch might predict a \texttt{thing} class, but the object detector detects no object at this location. In our current setup, the panoptic head will predict the most likely \texttt{stuff} class for these pixels. However, it might be beneficial to make an \texttt{unknown} prediction for these cases during inference. Hence, we also investigate the effect of such post-processing steps for our method.

\PAR{Joint Training:} During training, we combine four different loss functions. Our object detection branch is trained with focal loss $L_{\text{FL}}$ for classification and a smooth L1 loss $L_\text{reg}$ for box regression. The semantic segmentation branch uses a cross-entropy loss $L_\text{s}$ per pixel, and our instance center prediction branch uses a L1 loss $L_\text{i}$ per pixel. Our network is trained with a weighted combination of the loss terms: $L = \lambda_o (L_{\text{FL}} + L_{\text{reg}}) + \lambda_s L_\text{s} + \lambda_i L_\text{i}$. For the proposed \emph{smallest-first} and \emph{highest-confidence} policies, we train our network without the instance center prediction, but we add this branch for the \emph{closest-center} policy. We investigate different weighting schemes experimentally. We note that methods choosing weights automatically are not feasible for us as they are either too expensive, \eg, require multiple backward passes \cite{Chen_18_PMLR}, or are known to have optimization problems when using the Adam optimizer \cite{Kendall_18_CVPR}.

%-------------------------------------------------------------------------
\section{Experiments}

Our goal is to demonstrate that our network can serve as a simple and fast yet accurate method for panoptic segmentation in applications with strong run-time constraints. We show that our individual branches achieve expected accuracy and maintain this in a multi-task setting. We confirm that our panoptic head is capable of combining results from object detection and semantic segmentation to panoptic segmentation and thus, relax the requirement to use instance segmentation input. Moreover, we show that our panoptic head can handle conflicting predictions and overlaps. We exhaustively evaluate our proposed policies, different weightings, and post-processing steps.

\PAR{Dataset:} We perform all our experiments on the COCO dataset \cite{Lin_14_ECCV}. The dataset contains 118K training, 5K validation, and 20K test images with panoptic segmentation annotations for 80 \texttt{things} and 53 \texttt{stuff} categories. The large number of classes makes this dataset much more challenging under run-time constraints than datasets with only a dozen categories.

\PAR{Evaluation Criteria:} For single-task performance, we use mean Intersection-over-Union (mIoU) \cite{Everingham_15_IJCV} and Average Precision (AP) \cite{Lin_14_ECCV} for semantic segmentation and object detection, respectively. For panoptic segmentation, we use the standard PQ metric \cite{Kirillov_19_CVPR_PanSeg}. The PQ\textsuperscript{th} and PQ\textsuperscript{st} scores refer to the PQ scores for the \texttt{things} and \texttt{stuff} subset of the data, respectively. The metric is defined as follows:

\[
PQ = \frac{\sum_{(p,q)\in \text{TP}}\text{IoU}(p,q)}{|\text{TP}|+\frac{1}{2}|\text{FP}|+\frac{1}{2}|\text{FN}|}
\]

where TP, FP and FN correspond to the set of true positives, false positives and false negatives. Predicted segments that have an IoU of greater than 0.5 with a ground-truth segment belong to the TP set.

\PAR{Experimental Setup:} We use ImageNet pre-trained weights for our ResNet-50 encoder \cite{He_16_CVPR}. The FPN decoder and all branches are trained from scratch. We implement our model with PyTorch \cite{Paszke_17_NIPS} and train it on two Tesla V100 GPUs. In contrast to most related work that train with a batch size of 16, we use a batch size of 4. Hence, we freeze the BatchNorm layers \cite{Ioffe_15_ICML} in the encoder and omit fine-tuning their statistics. We train with Adam, a learning rate of 1e-5, a weight decay of 1e-4 and set $\beta_1=0.9$ and $\beta_2=0.99$. We limit the training to 14 epochs and decay the learning rate by a factor of 10 after the 7th and 10th epoch. All images are scaled such that the shorter side is at least 576px as long as the longer side has less than 864px. For training, we only apply left-right flipping as augmentation.

\subsection{Individual Components}

We validate our implementation of each branch by training them separately for their dedicated task. For brevity, we focus on object detection and instance center prediction. Our semantic segmentation branch comes with 26.9\% mIoU close to the reference score by \cite{Kirillov_19_CVPR_PanFPN} of 27.8\% mIoU on the COCO Stuff dataset with 92 classes and, thus, confirms that our network is indeed able to perform semantic segmentation. We validate our object detection branch by comparing to the official RetinaNet \cite{Lin_17_ICCV} scores on the COCO Detection dataset with 80 classes. The instance center prediction is evaluated on the COCO Panoptic dataset by recording the number of cases that \textit{intra-class} overlaps of ground-truth bounding boxes are resolved correctly.

\PAR{Object Detection:} We follow the proposed hyper-parameters by Lin \etal~\cite{Lin_17_ICCV} for our RetinaNet implementation. The reference implementation was trained with a batch size of 16 and SGD. Using the same ResNet-50 backbone at a similar input size, 
%Table~\ref{tab:results-retinanet} shows that we obtain slightly worse results, which we attribute to different training configurations and rescale policies.
we obtain slightly worse results with 33.1\% AP compared to 34.3\% AP reference score, which we attribute to different training configurations.

%\begin{table}
%	\centering
%	\input{tables/results-inst-center}
%	\caption{The instance center prediction is able to assign the correct ground-truth bounding box in 90.8\% of the intrac-class overlapping cases. The results are obtained on the COCO \textit{val2017} dataset.}
%	\label{tab:results-inst-center}
%\end{table}

\PAR{Instance Center Prediction:} In our proposed network, the instance center predictions are used to resolve id assignment for \textit{intra-class} overlaps. Hence, we evaluate the branch by tracking correct id assignments for overlaps with ground-truth bounding boxes.
%Table~\ref{tab:results-inst-center} shows that our branch can indeed be used for this task. We note that due to the 'crowd' annotations, many overlaps are not covered. Therefore, we also give qualitative examples in Figure~\ref{fig:instance-center-prediction}.
For 90.8\% of the pixels belonging to multiple boxes of the same class, the branch can assign the correct ground-truth bounding box. Hence, it can indeed be used for this task. We note that due to the 'crowd' annotations in COCO, many \textit{intra-class} overlaps are not considered for computing the final score, which limits our ability to showcase the strength of this branch. Therefore, we also show qualitative examples in Figure~\ref{fig:instance-center-prediction}.

\newcommand{\mycolsize}{0.31}
\newcommand{\mycolspace}{-10pt}

\begin{figure}
    \centering
    \begin{subfigure}[b]{\mycolsize\linewidth}
        \includegraphics[width=1.00\linewidth]{images/input/000000238866.jpg} \\ \vspace{\mycolspace}
        \includegraphics[width=1.00\linewidth]{images/input/000000370677.jpg} \\ \vspace{\mycolspace}
        \caption{Input}
    \end{subfigure}
    \begin{subfigure}[b]{\mycolsize\linewidth}
        \includegraphics[width=1.00\linewidth]{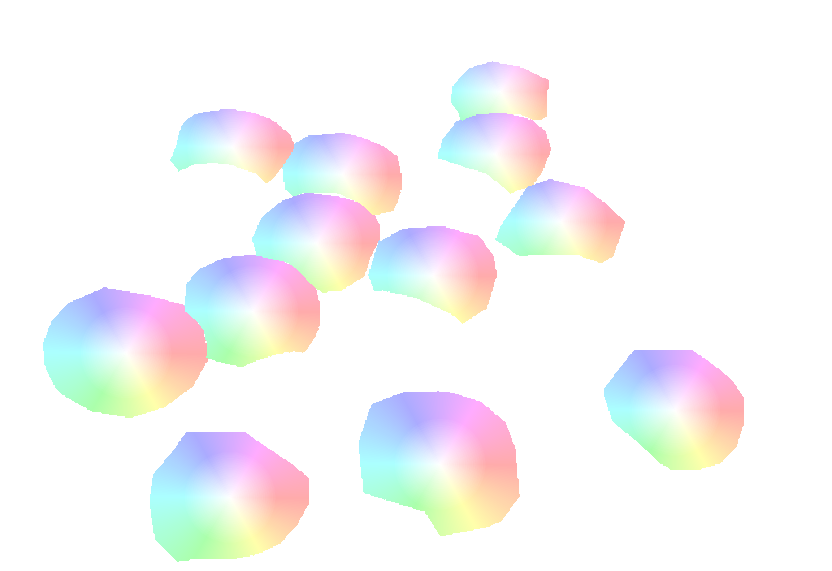} \\ \vspace{\mycolspace}
        \includegraphics[width=1.00\linewidth]{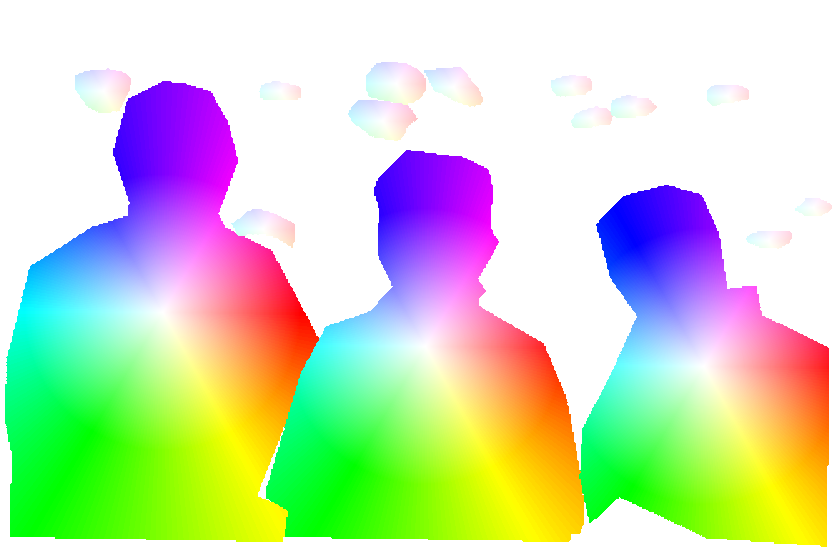} \\ \vspace{\mycolspace}
        \caption{Ground Truth}
    \end{subfigure}
    \begin{subfigure}[b]{\mycolsize\linewidth}
        \includegraphics[width=1.00\linewidth]{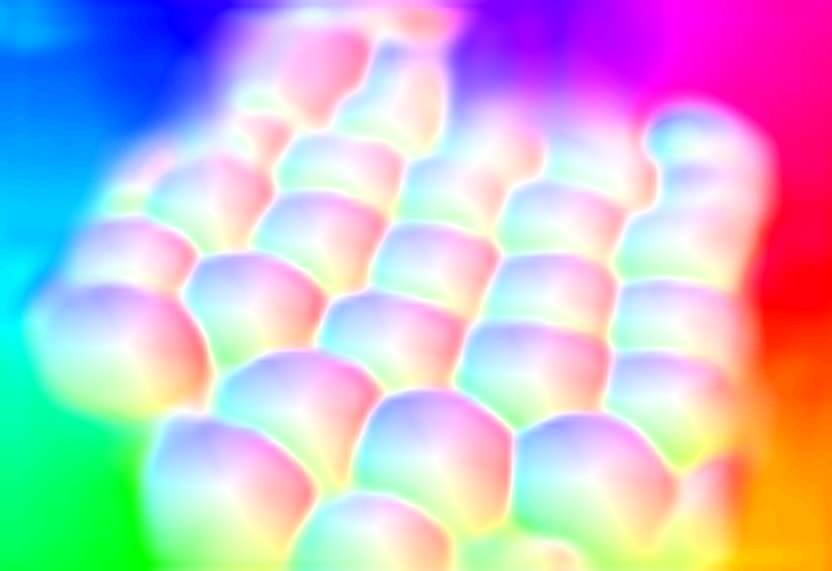} \\ \vspace{\mycolspace}
        \includegraphics[width=1.00\linewidth]{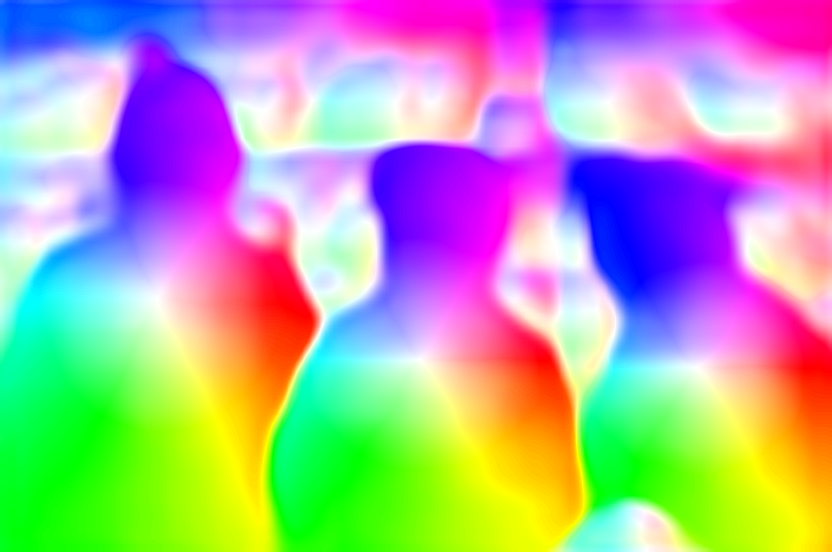} \\ \vspace{\mycolspace}
        \caption{Prediction}
    \end{subfigure}
    \caption{The results provide evidence that the instance center prediction branch can achieve smooth contours of overlapping instances of the same task. The visualization reflects the direction and magnitude of the offset vectors.}
	\label{fig:instance-center-prediction}
\end{figure}

\subsection{Panoptic Segmentation}

Having verified that our components work individually, we now investigate the performance of our panoptic head. Furthermore, we show the effect of the commonly applied \texttt{unknown} predictions. Finally, we examine the speed-accuracy trade-off and compare it to related work.

\begin{table}
	\centering
	\begin{tabular}{c|c||ccc|c|c}
\hline 
\multicolumn{7}{c}{\textbf{COCO Panoptic}}\tabularnewline
\hline 
\multicolumn{2}{c||}{Weights} & \multicolumn{3}{c|}{Pan. Seg.} & Seg. & Det.\tabularnewline
Sem. & Obj. & \textbf{PQ} & PQ\textsuperscript{th} & PQ\textsuperscript{st} & mIoU & AP\tabularnewline
\hline
\hline
1.00 & - & - & - & - & \textbf{50.0} & - \tabularnewline
- & 1.00 & - & - & - & - & \textbf{31.2} \tabularnewline
\hline
0.20 & 0.80 & 28.9 & 34.1 & 21.1 & 48.4 & \textbf{30.5}\tabularnewline
0.25 & 0.75 & 29.4 & 34.6 & 21.7 & 49.0 & \textbf{30.5}\tabularnewline
0.33 & 0.66 & \textbf{29.6} & \textbf{34.7} & 21.9 & 49.6 & 30.2\tabularnewline
\textbf{0.50} & \textbf{0.50} & \textbf{29.6} & 34.2 & 22.5 & 50.2 & 29.2\tabularnewline
0.66 & 0.33 & 29.4 & 33.6 & \textbf{23.0} & \textbf{50.4} & 27.7\tabularnewline
0.75 & 0.25 & 28.8 & 32.7 & 22.9 & 50.2 & 26.7\tabularnewline
0.80 & 0.20 & 28.4 & 32.2 & 22.8 & 50.2 & 25.9\tabularnewline
\end{tabular}

	\caption{We show the influence of the weighting on the single-task's and PQ scores on the COCO \textit{val2017} dataset. Overall, an equal weighting achieves the best balance for panoptic segmentation with the \emph{highest-confidence} policy.}
	\label{tab:two_branches_segdet_with_pq}
\end{table}

\PAR{Policies:} For evaluating our network, we already have to select a policy to resolve \textit{intra-class} overlaps in our panoptic head. We choose the \emph{highest-confidence} policy as this strategy is used in previous work to resolve overlaps in Mask R-CNN. Additionally, we apply a confidence threshold of 0.4 to the outputs of our detector. 
%We provide experimental validation of this threshold in the supplementary material. 
One challenge in multi-task training is setting the loss weights $\lambda_o,\lambda_s, \lambda_i$ such that the overall accuracy is maximized. Table~\ref{tab:two_branches_segdet_with_pq} shows the accuracy of our proposed network with the \emph{highest-confidence} policy on the COCO \textit{val2017} dataset. We note that the single-task models are trained on the same dataset with the same training configuration, while previously, we reported scores on different datasets. The scores provide evidence that equal weighting achieves the best balance between \texttt{things} and \texttt{stuff} for panoptic segmentation. Still, the results suggest that there is a range of weights that perform similarly well. Notably, the single-task performance for semantic segmentation is resilient to weight changes and even slightly benefits from the multi-task training. However, the detection task suffers from multi-task training, and its score highly depends on the weighting.

\begin{table}
	\centering
	\begin{tabular}{c||ccc|c|c}
\hline 
\multicolumn{6}{c}{\textbf{COCO Panoptic}}\tabularnewline
\hline
Policy & \textbf{PQ} & PQ\textsuperscript{th} & PQ\textsuperscript{st} & mIoU & AP\tabularnewline
\hline
\hline
HC & \textbf{29.6} & \textbf{34.2} & \textbf{22.5} & 50.2 & 29.2\tabularnewline
SF & 29.3 & 33.8 & \textbf{22.5} & 50.2 & 29.2\tabularnewline
\end{tabular}

	\caption{We evaluate the \emph{highest-confidence} (HC) and \emph{smallest-first} (SF) policy on the COCO \textit{val2017} dataset. The evidence supports the concept of resolving intra-class overlaps based on the confidence and not on the size.}
	\label{tab:results-hc-sf}
\end{table}

We can use the same trained networks to test the \emph{smallest-first} policy, as it only takes effect during inference. Similar to the \emph{highest-confidence} policy, an equal weighting performs best. 
%We provide an exhaustive analysis in the supplementary material. 
Table~\ref{tab:results-hc-sf} shows a comparison of both policies. The results indicate that the often applied overlap resolution by confidence score sorting is superior to size-based sorting.
%Thus, when using only two branches of our network, the \emph{highest-confidence} policy should be used.

\begin{figure}
    \centering
    \begin{subfigure}[b]{\mycolsize\linewidth}
        \includegraphics[width=1.00\linewidth]{images/input/000000002299.jpg} \\ \vspace{\mycolspace}
        \includegraphics[width=1.00\linewidth]{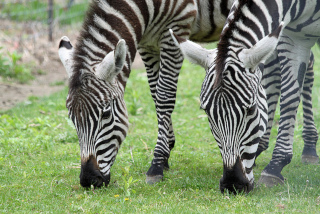} \\ \vspace{\mycolspace}
       	\includegraphics[width=1.00\linewidth]{images/input/000000054123.jpg} \\ \vspace{\mycolspace}
        \caption{Input}
    \end{subfigure}
    \begin{subfigure}[b]{\mycolsize\linewidth}
        \includegraphics[width=1.00\linewidth]{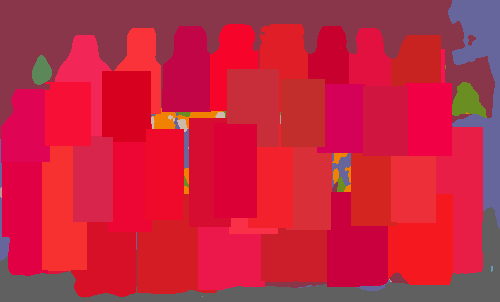} \\ \vspace{\mycolspace}
        \includegraphics[width=1.00\linewidth]{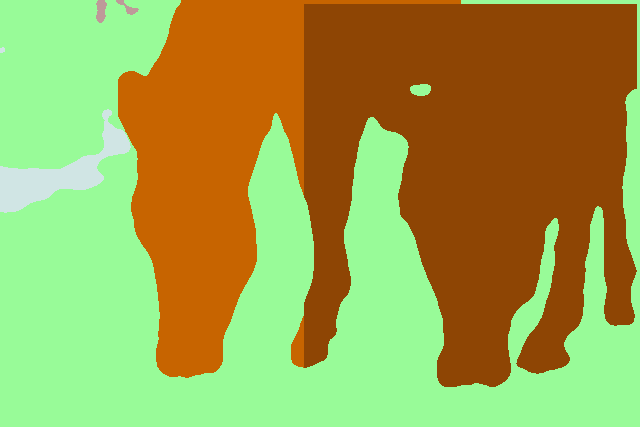} \\ \vspace{\mycolspace}
        \includegraphics[width=1.00\linewidth]{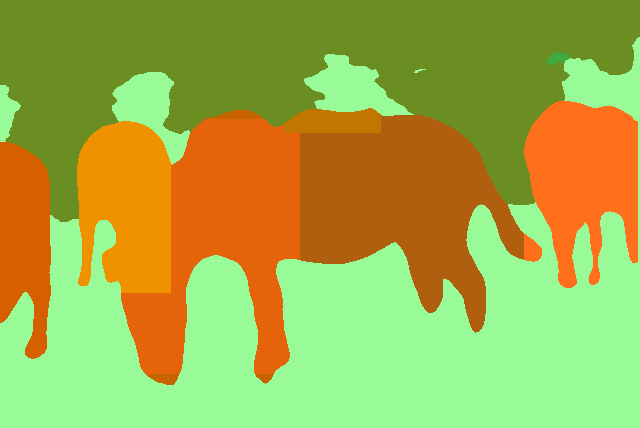} \\ \vspace{\mycolspace}
        \caption{Smallest-First}
    \end{subfigure}
    \begin{subfigure}[b]{\mycolsize\linewidth}
        \includegraphics[width=1.00\linewidth]{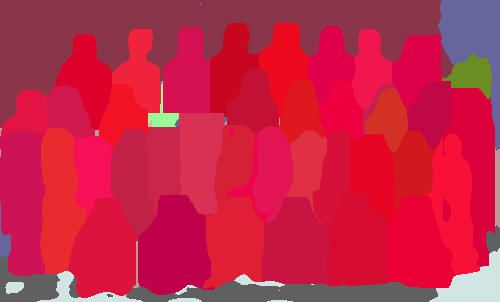} \\ \vspace{\mycolspace}
        \includegraphics[width=1.00\linewidth]{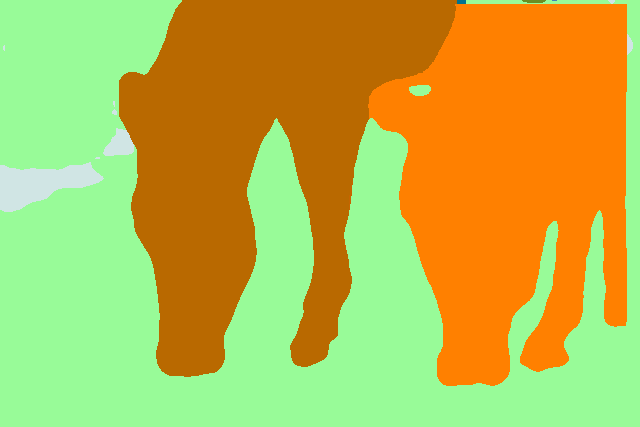} \\ \vspace{\mycolspace}
        \includegraphics[width=1.00\linewidth]{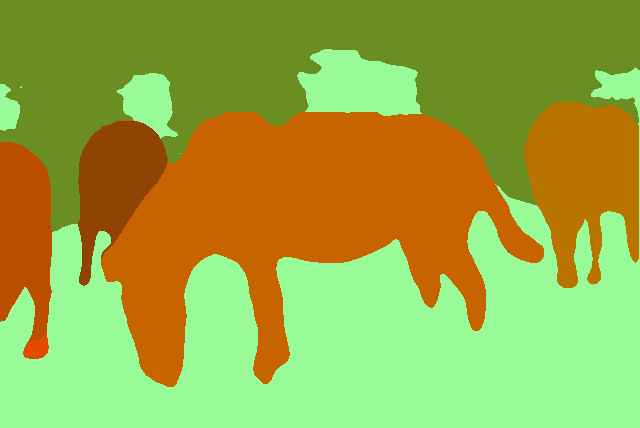} \\ \vspace{\mycolspace}
        \caption{Closest-Center}
    \end{subfigure}
    \caption{We compare the panoptic segmentation produced by our \emph{smallest-first} and \emph{closest-center} policy. The latter achieves much more visually plausible results.}
	\label{fig:closest-center}
\end{figure}

By the design of the \emph{highest-confidence} and \emph{smallest-first} policy, they lead to corner-shaped assignments of \textit{intra-class} overlaps. To overcome this, we introduced the third branch in combination with the \emph{closest-center} policy to achieve visually plausible contours. We confirm that this setup works to achieve panoptic segmentation and investigate different loss weightings in Table~\ref{tab:results-three-branches}. Since the initial loss for the instance center prediction is almost two order of magnitudes higher than for the other branches, we use a small weighting for the third branch. While the score increases only slightly, the qualitative comparison in Figure~\ref{fig:closest-center} shows the strength of our approach. We argue that the benefits of this approach are not well reflected in the COCO dataset. The advantages of the \emph{closest-center} policy play out in images that have a lot of overlapping instances of the same class. However, those regions are often not densely annotated but are labeled as crowd regions and excluded from the evaluation.

\begin{table}
	\centering
	\begin{tabular}{c|c||ccc|c|c}
\hline 
\multicolumn{7}{c}{\textbf{COCO Panoptic}}\tabularnewline
\hline 
\multicolumn{2}{c||}{Weights} & \multicolumn{3}{c|}{Pan. Seg.} & Seg. & Det.\tabularnewline
Sem./Obj. & Inst. & \textbf{PQ} & PQ\textsuperscript{th} & PQ\textsuperscript{st} & mIoU & AP\tabularnewline
\hline 
\hline 
1.0/- & - & - & - & - & \textbf{50.0} & -\tabularnewline
-/1.0 & - & - & - & - & - & \textbf{31.2}\tabularnewline
\hline 
0.5 & 0.001 & 29.7 & 34.5 & \textbf{22.6} & \textbf{50.3} & \textbf{29.5}\tabularnewline
0.5 & 0.005 & 29.8 & \textbf{34.8} & 22.3 & 49.9 & 29.3\tabularnewline
0.5 & 0.01  & \textbf{29.9} & \textbf{34.8} & 22.3 & 49.7 & 29.4\tabularnewline
\hline 
0.5 & GT & 31.4 & 37.3 & 22.5 & 50.2 & 29.2\tabularnewline
\end{tabular}

	\caption{We list the PQ scores of our final neural network under three different weightings and the \emph{closest-center} policy on the COCO \textit{val2017} split. Compared to the other policies, performance increases by 0.3--0.6\% PQ. We also show that better scores of the individual tasks do not correspond directly to better PQ scores and give an upper bound that can be achieved with ground-truth instance centers.}
	\label{tab:results-three-branches}
\end{table}

\begin{table}[ht]
	\centering
	\begin{tabular}{c|c|ccc}
\hline 
\multicolumn{5}{c}{\textbf{COCO Panoptic}}\tabularnewline
\hline 
Stuff Rem. & Unk. & \textbf{PQ} & PQ\textsuperscript{th} & PQ\textsuperscript{st}\tabularnewline
\hline
\hline
- & - & 29.9 & 34.8 & 22.5\tabularnewline
$\checkmark$ & - & 32.2 \textit{\footnotesize(+2.3)} & 34.8 & 28.4 \textit{\footnotesize(+5.9)}\tabularnewline
$\checkmark$ & $\checkmark$ & 32.4 \textit{\footnotesize(+2.5)} & 34.8 & 28.6 \textit{\footnotesize(+6.1)}\tabularnewline
\end{tabular}

	\caption{We experiment with \texttt{unknown} predictions (Unk.) and small \texttt{stuff} removal (Stuff Rem.). We evaluate our network with the \emph{closest-center} policy on COCO \textit{val2017}. Using a threshold for \texttt{stuff} regions and adding \texttt{unknown} predictions increases the score by 2.5\% PQ.}
	\label{tab:ablation-unknown}
\end{table}

\PAR{Unknown Predictions:} We investigate the effect of using \texttt{unknown} predictions and removing small \texttt{stuff} regions. We set the threshold to 4096 pixels for \texttt{stuff} regions, as is best practice \cite{Kirillov_19_CVPR_PanFPN}. Table~\ref{tab:ablation-unknown} lists the scores of our \emph{closest-center} policy with these post-processing steps. Remarkably, the removal of small \texttt{stuff} regions leads to an increase of 2.3\% PQ. Since several related methods use the same removal step without any ablations, we hypothesize that the used semantic segmentation branches suffer from the same issue, which is over-segmentation of \texttt{stuff} regions.

\begin{table}
	\centering
	\begin{tabular}{c|c||ccc}
\hline 
\multicolumn{5}{c}{\textbf{COCO Panoptic}}\tabularnewline
\hline
Name & Backbone & \textbf{PQ} & PQ\textsuperscript{th} & PQ\textsuperscript{st}\tabularnewline
\hline 
\hline 
PanFPN \cite{Kirillov_19_CVPR_PanFPN}& ResNet-50 & 39.0 & 45.9 & 28.7\tabularnewline
OANet \cite{Liu_19_CVPR}& ResNet-50 & 39.0 & 48.3 & 24.9\tabularnewline
AUNet \cite{Li_19_CVPR}& ResNet-50 & 39.6 & \textbf{49.1} & 25.2\tabularnewline
UPSNet \cite{Xiong_19_CVPR}& ResNet-50 & \textbf{42.5} & 48.5 & \textbf{33.4}\tabularnewline
%\hline 
%JSIS$^{\dag}$ \cite{DeGeus_19_arXiv}& ResNet-50 & 26.9 & 29.3 & 23.3\tabularnewline
%OCFusion$^{\dag}$ \cite{Lazarow_19_arXiv}& ResNet-50 & 41.0 & 49.0 & 29.0\tabularnewline
AdaptIS$$ \cite{Sofiiuk_19_arXiv}& ResNet-50 & 35.9 & 40.3 & 29.3\tabularnewline
\hline 
\hline
DeeperLab$^{\dag}$ \cite{Yang_19_arXiv}& LWMNV2 & 24.1 & - & -\tabularnewline
DeeperLab$^{\dag}$ \cite{Yang_19_arXiv}& WMNV2 & 27.9 & - & -\tabularnewline
DeeperLab$^{\dag}$ \cite{Yang_19_arXiv}& Xception-71 & 33.8 & - & -\tabularnewline
\hline 
Ours (SF)& ResNet-50 & 31.8 & 33.8 & 28.9\tabularnewline
Ours (HC)& ResNet-50 & 32.1 & 34.2 & 28.8\tabularnewline
Ours (CC)& ResNet-50 & 32.4 & 34.8 & 28.6\tabularnewline
\end{tabular}

	\caption{We report the performance on COCO \textit{val2017} of related work using two-stage networks, the concurrent ($\dag$) work on single-shot methods, and our approach. We limit this comparison to the ResNet-50 for two-stage methods. SF, HC, and CC refer to our policies.}
	\label{tab:comparison-accuracy}
\end{table}

\PAR{Final Results:} We use \texttt{unknown} predictions to report our final scores as is common practice. Table~\ref{tab:comparison-accuracy} provides results from published two-stage methods and the concurrent single-shot approach DeeperLab \cite{Yang_19_arXiv}. In contrast to some other approaches, our scores are not obtained with test-time tricks (multi-scale, l-r flipping) or other backbone optimizations, \eg, deformable convolutions \cite{Dai_17_ICCV}, that are known to improve scores \cite{Xiong_19_CVPR, Lazarow_19_arXiv}, due to run-time considerations. Moreover, all approaches achieve consistently better scores with deeper and wider backbones. For a fair comparison of our method, we report scores for a ResNet-50 encoder. 
%Notably, the accuracy gap between our single-shot method and two-stage approaches is mostly less than ten percentage points. 
Notably, our network performs more accurately than two DeeperLab variants. The configuration using the Xception-71 backbone~\cite{Chollet_17_CVPR} performs slightly better but uses also a far deeper backbone. The Xception-71 encoder has 42.1M parameters, while our encoder has only 23.6M parameters.

\begin{table*}
	\centering
	\begin{tabular}{c|c|c||ccc|c}
\hline 
\multicolumn{7}{c}{\textbf{COCO Panoptic Test}}\tabularnewline
\hline 
Name & Backbone & Input size & \textbf{PQ} & PQ\textsuperscript{th} & PQ\textsuperscript{st} & FPS\tabularnewline
\hline 
\hline 
Ours (SF) & ResNet-50 & $576\times[576,864]$ & 32.1 & 34.3 & 28.8 & 23.9\tabularnewline
Ours (HC) & ResNet-50 & $576\times[576,864]$ & 32.2 & 34.5 & 28.8 & 23.9\tabularnewline
Ours (CC) & ResNet-50 & $576\times[576,864]$ & 32.6 & 35.0 & 29.0 & 23.5\tabularnewline
%Ours (CC) & ResNet-50 & $576\times[576,864]$ & 32.6 & 35.0 & 29.0 & 21.3\tabularnewline
\hline 
DeeperLab \cite{Yang_19_arXiv} & LWMNV2 & $321\times321$ & 18.0 & 18.5 & 17.2 & 26.6\tabularnewline
DeeperLab \cite{Yang_19_arXiv}& LWMNV2 & $641\times641$ & 24.5 & 26.9 & 20.9 & 13.7\tabularnewline
DeeperLab \cite{Yang_19_arXiv}& WMNV2 & $641\times641$ & 28.1 & 30.8 & 24.1 & 12.0\tabularnewline
DeeperLab \cite{Yang_19_arXiv}& Xcept.-71 & $641\times641$ & \textbf{34.3} & \textbf{37.5} & \textbf{29.6} & 8.4\tabularnewline
\end{tabular}

	\caption{We compare our method to DeeperLab on COCO \textit{test-dev}. The FPS incorporates the time from input to output. Under similar input conditions, we outperform DeeperLab in all cases regarding the run-time and in two out of three cases regarding the score. SF, HC, and CC refer to our \emph{smallest-first}, \emph{highest-confidence} and \emph{closest-center} policy.}
	\label{tab:comparison-test}
\end{table*}

\begin{figure}
	\centering
	\includegraphics[width=1.00\linewidth]{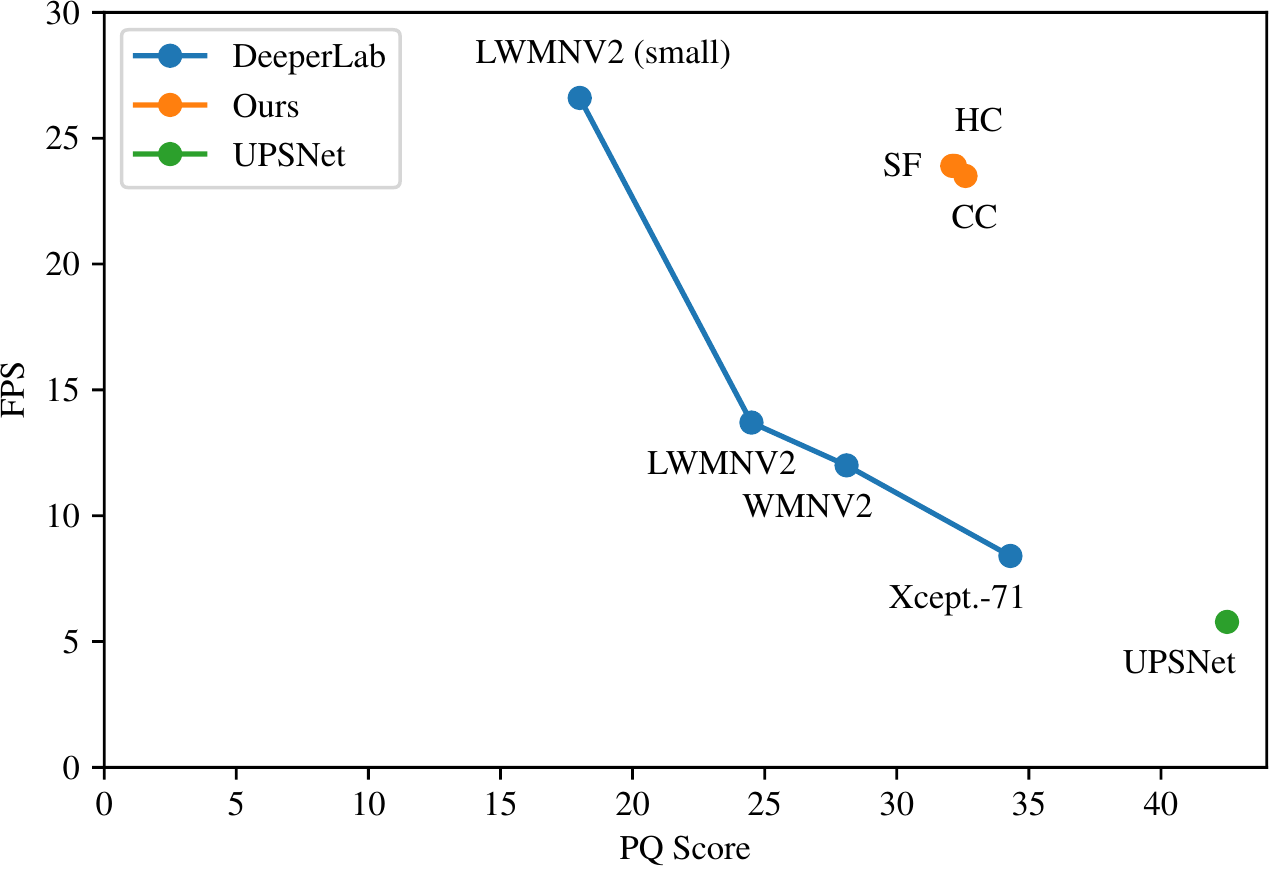}
	\caption{We visualize the trade-off between accuracy and speed. DeeperLab trades a slightly better score for a low run-time that is far from video-rate. The plot shows that our network achieves a better trade-off and therefore, is beneficial for real-world applications.}
	\label{fig:trade-off}
\end{figure}

\begin{figure}
    \centering
    \begin{subfigure}[b]{\mycolsize\linewidth}
        \includegraphics[width=1.00\linewidth]{images/input/000000058705.jpg} \\ \vspace{\mycolspace}
        \includegraphics[width=1.00\linewidth]{images/input/000000060770.jpg} \\ \vspace{\mycolspace}
        \includegraphics[width=1.00\linewidth]{images/input/000000235857.jpg} \\ \vspace{\mycolspace}
        \includegraphics[width=1.00\linewidth]{images/input/000000449579.jpg} \\ \vspace{\mycolspace}
        \caption{Input}
    \end{subfigure}
    \begin{subfigure}[b]{\mycolsize\linewidth}
        \includegraphics[width=1.00\linewidth]{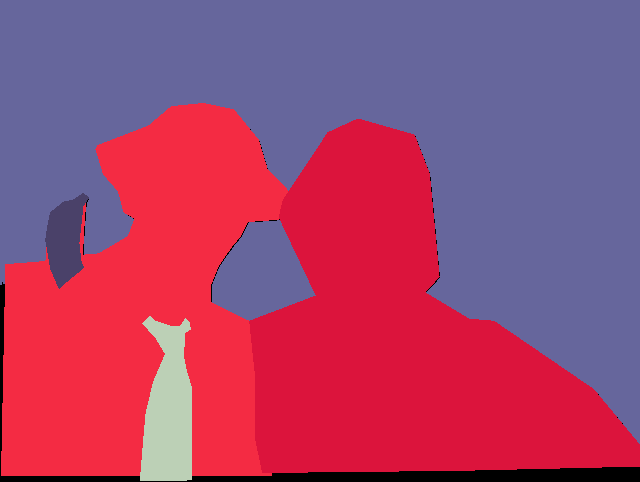} \\ \vspace{\mycolspace}
        \includegraphics[width=1.00\linewidth]{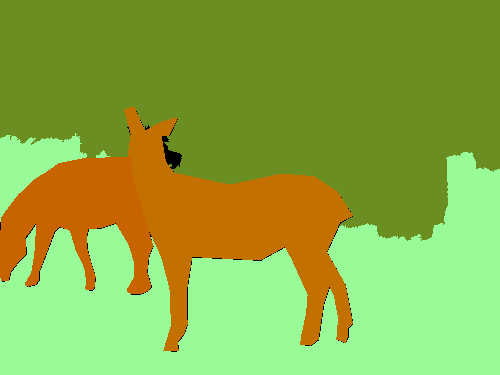} \\ \vspace{\mycolspace}
        \includegraphics[width=1.00\linewidth]{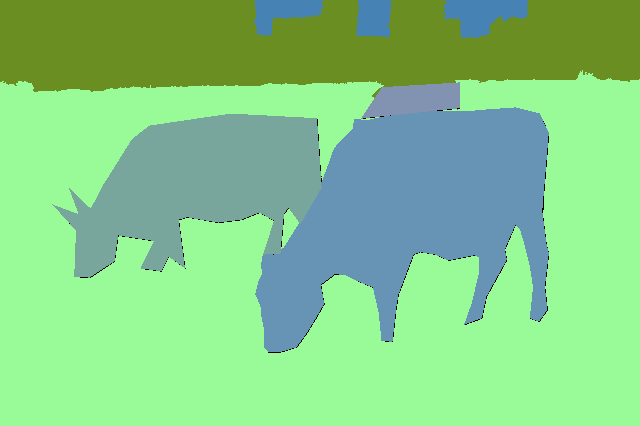} \\ \vspace{\mycolspace}
        \includegraphics[width=1.00\linewidth]{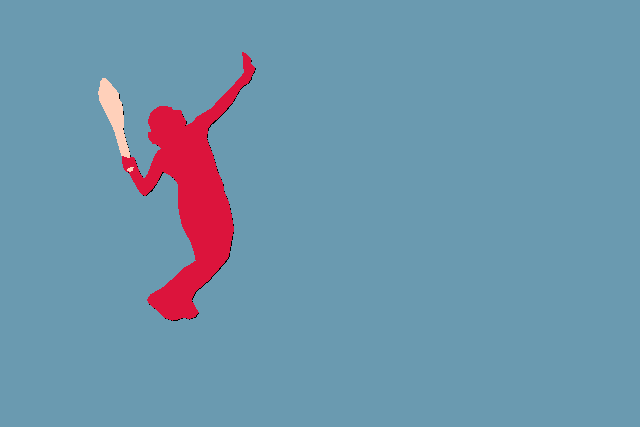} \\ \vspace{\mycolspace}
        \caption{Ground Truth}
    \end{subfigure}
    \begin{subfigure}[b]{\mycolsize\linewidth}
        \includegraphics[width=1.00\linewidth]{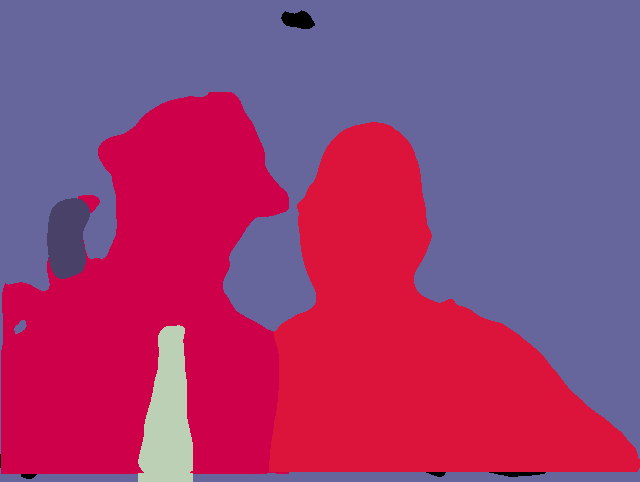} \\ \vspace{\mycolspace}
        \includegraphics[width=1.00\linewidth]{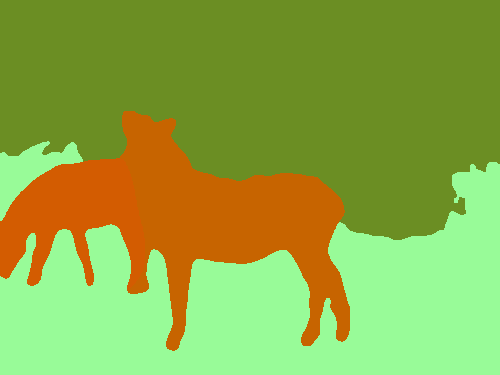} \\ \vspace{\mycolspace}
        \includegraphics[width=1.00\linewidth]{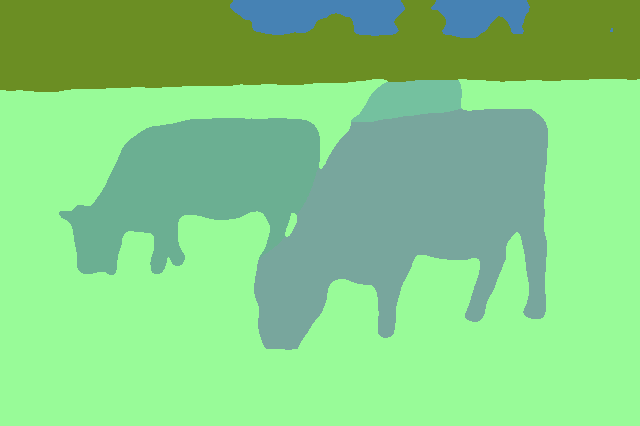} \\ \vspace{\mycolspace}
        \includegraphics[width=1.00\linewidth]{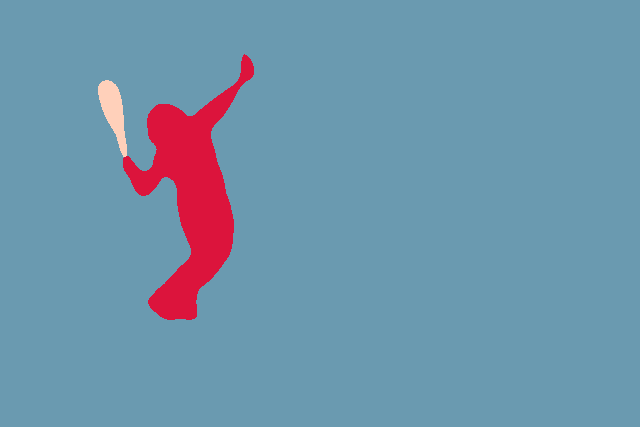} \\ \vspace{\mycolspace}
        \caption{Prediction}
    \end{subfigure}
    \caption{Qualitative results from our final network with the \emph{closest-center} policy on the COCO \textit{val2017} dataset.}
	\label{fig:final-results}
\end{figure}

The large Xception-71 is also the reason the frame rate of the most accurate DeeperLab is still far from video frame rate. We report the accuracy and FPS of the single-shot methods in Table~\ref{tab:comparison-test}. All run-times from our method and DeeperLab are measured on a Tesla-V100-SXM2 GPU. We calculate a robust FPS score by taking the median FPS of 6 runs over the whole COCO validation set. For DeeperLab, we use the official time measurements but include the timings from intermediate results to the final panoptic segmentation. In the case of DeeperLab, the intermediate results are heatmaps and semantic segmentation. However, we argue the FPS must measure the time from input to output to fully cover the run-time to obtain panoptic segmentation. Moreover, all fully convolutional networks run faster (and usually perform worse) on smaller input as fewer computations are necessary. Hence, we compare methods with similar input conditions for a fair evaluation. All our networks run significantly faster under these conditions than the competitor DeeperLab. With 23.5 to 23.9 FPS, we almost achieve video frame rate. Moreover, our method is 4 times faster than state-of-the-art two-stage approach UPSNet \cite{Xiong_19_CVPR}. Using the official implementation with a ResNet-50, UPSNet achieves 5.8 FPS. Figure~\ref{fig:trade-off} 
%shows the speed-accuracy trade-off and 
%provides evidence 
shows that our network achieves a significantly better speed-accuracy trade-off, making our approach more broadly applicable. We show qualitative results in Figure~\ref{fig:final-results}.

%-------------------------------------------------------------------------
\section{Conclusion}

In this paper, we present a novel end-to-end single-shot panoptic segmentation approach that achieves remarkable accuracy, while setting a new state-of-the-art run-time performance at almost video frame rate. Our panoptic head does not simply merge outputs of sub-networks but produces instance-aware panoptic logits. In contrast to previous work, our proposed method does not rely on instance segmentation, but can be used with detectors widely used in the robotics community. Still, our panoptic head can resolve \textit{inter-} and \textit{intra-class} overlaps by combining semantic segmentation, object detection and instance center prediction. This panoptic head is also able to correct errors in each of its inputs before creating the final panoptic segmentation result, which is superior to merge-by-copy methods.

\PAR{Acknowledgements:} This  project  has  been  funded,  in  parts, by ERC Consolidator Grant DeeViSe (ERC-2017-COG-773161).  Simulations were performed with computing resources granted by the RWTH Aachen University under project rwth0431. 
We would like to thank Tobias Fischer for helpful discussions.

\bibliographystyle{IEEEtran}
\bibliography{IEEEabrv,weber-iros-2020}
%}

\end{document}